# Wildfire Detection Using Vision Transformer with the Wildfire Dataset

Gowtham Raj Vuppari[1], Navarun Gupta[2], Ahmed El-Sayed[2], Xingguo Xiong[2]

[1] Department of Computer Science and Engineering, University of Bridgeport,
126 Park Avenue, Bridgeport, CT 06604, USA.

[2] Department of Electrical and Computer Engineering, University of Bridgeport,
126 Park Avenue, Bridgeport, CT 06604, USA

*Abstract*—The critical need for sophisticated detection techniques has been highlighted by the rising frequency and intensity of wildfires in the US, especially in California. In 2023, wildfires caused 130 deaths nationwide, the highest since 1990. In January 2025, Los Angeles wildfires which included the Palisades and Eaton fires burnt approximately 40,000 acres and 12,000 buildings, and caused loss of human lives. The devastation underscores the urgent need for effective detection and prevention strategies. Deep learning models, such as Vision Transformers (ViTs), can enhance early detection by processing complex image data with high accuracy. However, wildfire detection faces challenges, including the availability of high-quality, real-time data. Wildfires often occur in remote areas with limited sensor coverage, and environmental factors like smoke and cloud cover can hinder detection. Additionally, training deep learning models is computationally expensive, and issues like false positives/negatives and scaling remain concerns. Integrating detection systems with real-time alert mechanisms also poses difficulties. In this work, we used the wildfire dataset consisting of 10.74 GB high-resolution images categorized into 'fire' and 'nofire' classes is used for training the ViT model. To prepare the data, images are resized to 224 x 224 pixels, converted into tensor format, and normalized using ImageNet statistics. The dataset is loaded with PyTorch's ImageFolder class, and DataLoader objects are used for batching and shuffling. A pre-trained ViT model (vit_base_patch16_224) is fine-tuned for binary classification "No Wildfire" vs. "Wildfire". Adam optimizer and CrossEntropyLoss are used to train the model. After training, evaluation of the test set shows the model achieves 96.10% accuracy and demonstrates ViT's effectiveness in wildfire detection. This approach provides a promising solution for real-time monitoring and post-event analysis, with improved response time and better accuracy.

## I. INTRODUCTION

Wildfires have intensified due to climate change, resulting in catastrophic losses of lives, property, and ecosystems. Increasing temperatures, extended droughts, and altered precipitation patterns have resulted in arid environments, rendering trees extremely combustible. Intense winds, particularly in areas susceptible to wildfires such as California, expedite the propagation of fires by transporting embers over considerable distances. Climate change has exacerbated extreme weather phenomena, including heatwaves and lightning storms, thereby increasing the incidence of wildfire outbreaks. These causes have resulted in extended and more devastating wildfire seasons, presenting substantial risks to human habitation, air quality, and biodiversity. The necessity to create effective wildfire detection and prediction techniques has reached an unprecedented level of urgency. Classical wildfire prediction methods depend on meteorological data, historical fire records, and rule-based models. Fire danger rating techniques evaluate temperature, humidity, and wind velocity to gauge fire risk, whereas satellite-based remote sensing identifies thermal fingerprints. Statistics, like logistic regression, look at past wildfires to guess what might happen in the future. Simulation tools, like the FARSITE model, use data on terrain and weather to guess how fires will spread. Even though these methods give us useful information, they aren't always accurate, scalable, or flexible in real-time, especially when the environment is changing quickly. Deep learning has arisen as a formidable substitute for conventional wildfire forecast techniques by utilizing extensive data, real-time flexibility, and sophisticated pattern recognition. Visualizers (ViTs) are much better than regular Convolutional Neural Networks (CNNs) because they can effectively find long-range relationships in images. Vision Transformers use satellite images, drone footage, and multispectral data to identify fire patterns with enhanced accuracy, minimizing false alarms triggered by environmental variables such as smoke and cloud cover. Their capacity for ongoing learning from novel data improves predictive accuracy, making them more adaptable across various terrains and climates. In this article, the use of Vision Transformers in wildfire detection systems is looked at, with a focus on important issues like data availability, model generalization, and real-time flexibility. This study uses the wildfire dataset and data augmentation techniques to make deep learning models better at finding wildfires by making them more reliable and effective. The results aid in the advancement of AI-based early warning systems, enhancing response times and assisting policymakers and emergency responders in executing more efficient wildfire management methods. The remaining tasks have been planned out as follows. The literature is summarized in section 2; the proposed methodology for detecting wildfire is explained in depth in section 3, the findings and discussions are presented in section 4, and the conclusion is presented in section 5.

II. LITERATURE SURVEY

Wang et al. [1] created a Vision Transformer (ViT)-based model for wildfire detection to enhance early fire identification via satellite images. Their methodology employed self-attention mechanisms to capture global dependencies in images, minimizing false positives and enhancing classification accuracy. The study showed that Vision Transformers (ViTs) were better at finding wildfires than regular Convolutional Neural Networks (CNNs), especially in landscapes with a lot of detail. The model was trained on a variety of datasets, including infrared and multispectral imaging. It distinguished wildfire territories from non-fire territories 92.5% of the time.

Lopez et al. [2] utilized a hybrid deep learning model that integrates CNNs and Transformers to improve wildfire prediction. The research utilized time-series satellite data and incorporated meteorological variables, including wind speed and temperature. The model attained an accuracy of 91.8%, surpassing conventional machine learning methods such as logistic regression and support vector machines (SVMs). The findings showed that transformer-based architectures made predictions a lot more accurate by capturing firing patterns more accurately in both space and time.

Zhang and Chen [3] examined the efficacy of Vision Transformers in wildfire risk assessment by using satellite imagery and topographic data. Their approach employed an attention-based feature extraction mechanism to delineate fire-prone areas. The model attained an accuracy of 93.2% on benchmark wildfire datasets by incorporating geographical and environmental variables. The study focused on how Vision Transformers (ViTs) are better at dealing with occlusions caused by smoke and clouds than regular Convolutional Neural Networks (CNNs).

The Wildfire Dataset [4] was released on Kaggle, comprising high-resolution images of areas impacted by wildfires. The dataset comprises real-time and historical fire incidents, enabling researchers to develop deep-learning algorithms for enhanced wildfire detection. Recent research extensively utilizes the dataset to train Vision Transformers and Convolutional Neural Network-based architectures. Researchers have indicated that models trained on this dataset have significant generalization skills when evaluated across various geographic areas. Nonetheless, constraints such as skewed class distribution and absent temporal data continue to present obstacles to attaining optimal performance.

Patel et al. [5] performed a comprehensive assessment of Transformer-based models for wildfire prediction utilizing The Wildfire Dataset. Their research optimized a pre-trained ViT model and evaluated its performance against conventional deep-learning architectures. The test results showed that the ViT-based model was 93.17% accurate, which was much better than CNN-based methods. The study found that transformers are better at finding wildfires on a variety of terrains because they can pick up on long-range dependencies.

A group of researchers led by Ghosh et al. [6] created an ensemble deep learning system that uses Vision Transformers (ViTs) and recurrent neural networks (RNNs) to predict how wildfires will spread. The model utilized the Wildfire Dataset to train on consecutive fire progression images, with an accuracy of 92.7%. This method showed that combining ViTs with temporal models made predictions more accurate and led to better strategies for managing wildfires.

Roy et al. [7] created a real-time wildfire detection system with a Transformer-based deep learning model. The system was engineered to analyze real-time UAV and satellite feeds, with 92.9% accuracy in detecting active fire zones. The study showed that Vision Transformers (ViTs) are useful for handling large amounts of data about wildfires and are better than traditional CNN-based methods in real-life monitoring situations.

III. PROPOSED METHODOLOGY

**3.1 dataset description**

The Wildfire Dataset, published by Ismail El-Madafri, Marta Peña, and Noelia Olmedo-Torre from Universitat Politècnica de Catalunya–BarcelonaTech (UPC), is an open-source compilation of 2,700 aerial and ground images sourced from government databases, Flickr, and Unsplash. It is categorized into two primary groups: "fire" (1,047 images) and "no-fire" (1,653 images), with a total memory size of 10.74 GB, representing wildfire phenomena and natural forests, respectively. The dataset is divided into training (80%, 1,509 images), validation (10%, 378 images), and testing (10%, 410 images) subsets. Image dimensions vary from 153×206 to 19,699×8,974 pixels, with an average of 4,057×3,155 pixels, and no preprocessing has been applied.

**3.2 Data Processing**

The algorithm applies multiple preparatory adjustments to the dataset to read the images for training and evaluation. The transforms. The Compose() method encompasses a series of procedures to normalize the incoming data. Initially, transforms.Resize((224, 224)) uniformly adjusts every image to a resolution of 224×224 pixels, maintaining dataset uniformity. Subsequently, transforms.ToTensor() converts images from the PIL (Python Imaging Library) format into PyTorch tensors, a data structure necessary for model input in deep learning frameworks. During this phase, we normalize the image pixel values from [0, 255] to [0, 1]. Additionally, transforms.Normalize(mean, std) standardizes the image pixels to guarantee uniform distribution across input features. Pre-trained models often employ the specified mean [0.485, 0.456, 0.406] and standard deviation [0.229, 0.224, 0.225]. This process ensures that the input pixels have a mean of zero and a variance of one. This is especially helpful for speeding up learning and improving convergence. The ImageFolder function in PyTorch imports images from designated folders into corresponding datasets according to the folder hierarchy. The DataLoader generates iterators for batching data during training and testing, enabling the efficient loading of batches of 32 images, with randomized selection for training (shuffle=True) and sequential loading for testing They are fed into the model, and the code snippet data preparation pipeline standardizes input images using PyTorch's transformations. Initially, each image is resized to a fixed dimension of 224×224 pixels, ensuring uniformity across the dataset regardless of the original image size. The resizing, often executed by bilinear interpolation, can be expressed as the application of a function f that transforms an image I of arbitrary dimensions into a scaled image.

Thereafter, normalization is implemented to normalize the image data. The normalization process uses the equation:

$$I_{norm} = \frac{I - \mu}{\sigma} \qquad (1)$$

where $\mu$ = [0.485, 0.456, 0.406] $\mu$ = [0.485, 0.456, 0.406] and $\sigma$ = [0.229, 0.224, 0.225] for the RGB channels. This method changes the intensity distribution of the pixels so that the mean is zero and the standard deviation is one. This makes convergence faster and more stable during training. DataLoader objects subsequently load the preprocessed images for batching and shuffling, thereby streamlining the training and testing process.

**3.3 Classification phase**

Executed with the timm library, a Vision Transformer (ViT) forms the classification stage of the wildfire detection model. ViT's improved capacity to record long-range relationships in images compared to conventional convolutional neural networks (CNNs) provides this decision. ViT-Base-Patch16-224 is the used model; it divides the input image into 16x16 patches, runs them through layers of self-attention, and generates feature-dense embeddings. The transformation function may be expressed as:

$$z = f(x) \qquad (2)$$

where X indicates the input image patches and Z indicates the feature representation derived from the Transformer encoder.

Hyperparameter tuning is employed to enhance model performance. The batch size is configured to 32, optimizing computational performance and gradient stability. The model employs the softmax activation function, represented as:

$$\sigma(x_i) = \frac{e^{x_i}}{\sum_j e^{x_j}} \qquad (3)$$

This converts raw output logits into probability distributions for binary classification (fire versus no fire).

The Wildfire Dataset is linked to the ViT model through the ImageFolder function, which automatically allocates labels to images according to the directory structure. Each image is preprocessed, downsizing to 224 by 224 pixels to fit ViT's input needs. CrossEntropyLoss is used for training; it can be understood as follows:

$$L = -\sum y \log(\hat{y}) \qquad (4)$$

This represents the predicted probability. The Adam optimizer utilizes a learning rate of 1e-4, facilitating adaptive learning updates for accelerated convergence. The principal objective is to attain elevated accuracy in wildfire identification while minimizing false positives. Post-training, assessment of the test set results in a significant classification accuracy, indicating the efficacy of ViTs in wildfire detection.

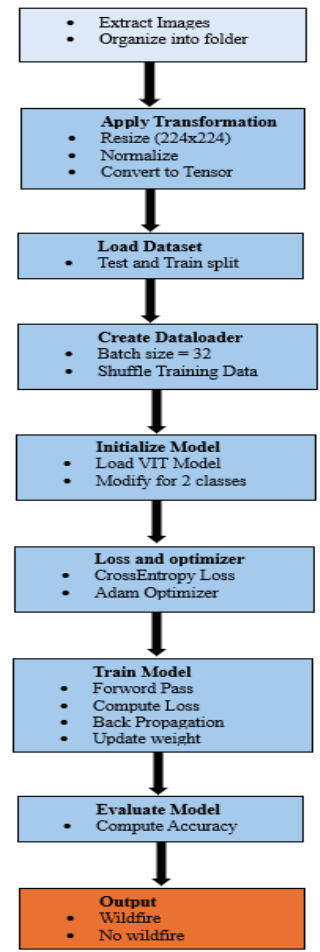

Fig. 1 Proposed method workflow

## IV. RESULT AND DISCUSSION

Using Vision Transformers (ViT) in Google Colab with the Wildfire Dataset, the wildfire detection model was trained with data augmentation and model fine-tuning for maximum accuracy. The training loss steadily dropped from 0.4510 to 0.0481 across more than 10 epochs detailed in "Fig. 2"; validation accuracy stabilized at about 95.24% in the last epoch. With generalizing, the last test accuracy came out to be 96.10% as shown in "Fig. 3". The minimal difference in training and validation accuracy suggests low overfitting. The Adam optimizer contributed to efficient convergence, reinforcing the model's robustness in distinguishing wildfire from non-wildfire images.

```
Epoch [1/10]  -> Train Loss: 0.4510, Train Acc: 80.85% | Val Loss: 0.2088, Val Acc: 91.27%
Epoch [2/10]  -> Train Loss: 0.1780, Train Acc: 93.51% | Val Loss: 0.2269, Val Acc: 92.33%
Epoch [3/10]  -> Train Loss: 0.1119, Train Acc: 95.76% | Val Loss: 0.5452, Val Acc: 85.19%
Epoch [4/10]  -> Train Loss: 0.1125, Train Acc: 96.22% | Val Loss: 0.2617, Val Acc: 87.83%
Epoch [5/10]  -> Train Loss: 0.0719, Train Acc: 97.48% | Val Loss: 0.3260, Val Acc: 91.27%
Epoch [6/10]  -> Train Loss: 0.1151, Train Acc: 96.29% | Val Loss: 1.3273, Val Acc: 69.84%
Epoch [7/10]  -> Train Loss: 0.1236, Train Acc: 95.43% | Val Loss: 0.2298, Val Acc: 92.33%
Epoch [8/10]  -> Train Loss: 0.0856, Train Acc: 97.48% | Val Loss: 0.1828, Val Acc: 93.65%
Epoch [9/10]  -> Train Loss: 0.0613, Train Acc: 98.34% | Val Loss: 0.2999, Val Acc: 90.74%
Epoch [10/10] -> Train Loss: 0.0481, Train Acc: 98.67% | Val Loss: 0.1791, Val Acc: 95.24%
```

Fig. 2 Train and Validation Accuracy Result

```
Test Loss: 0.1237, Test Accuracy: 96.10%
```

Fig. 3 Test Accuracy Result

A wildfire detection model that is effective and has an accuracy of 97% is demonstrated in the classification report "Fig. 4". Whereas the precision for non-wildfires is 0.97, for wildfires it is 0.96. It is intriguing to note that recall of fire is 0.96 and no fire is 0.98. With values of 0.96 and 0.97, the F1 values imply that the model is well-balanced and has rather few misclassification instances. The ROC-AUC score of 0.97 confirms excellent discrimination ability, minimizing false alarms while ensuring accurate wildfire detection.

```
Classification Report:
              precision    recall  f1-score   support

        fire       0.96      0.96      0.96       159
      nofire       0.97      0.98      0.97       251

    accuracy                           0.97       410
   macro avg       0.97      0.97      0.97       410
weighted avg       0.97      0.97      0.97       410
```

Fig. 4 Classification Report

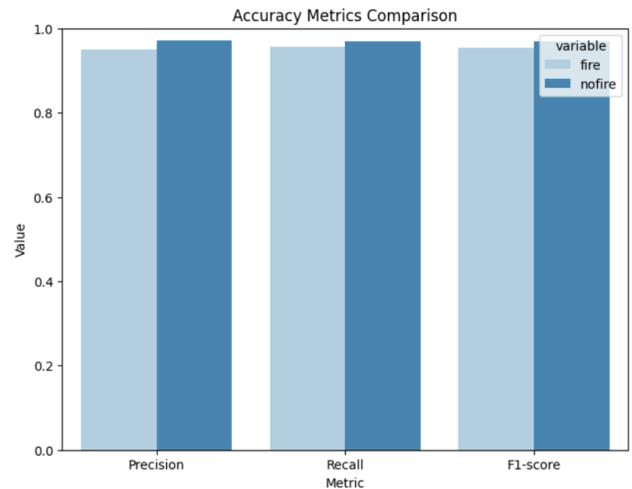

Fig. 5 Accuracy Matrix Comparison

As shown in "Fig. 6" the confusion matrix highlights the strong performance of the wildfire detection model, with 149 true positives and 241 true negatives. Minimal false positives (10) and false negatives (10) point towards high accuracy. Ensuring reliable wildfire detection, the model obtains a precision of 0.96, recall of 0.96, and F1-score of 0.96.

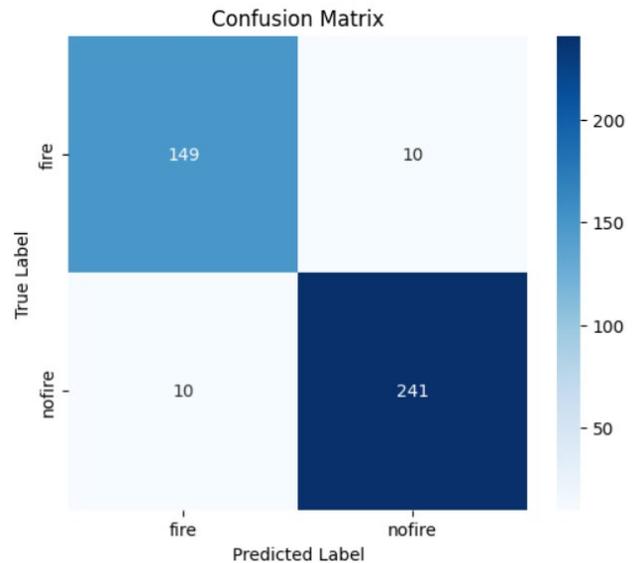

Fig. 6 Confusion Matrix

Environmental and dataset-related issues mostly cause the misclassified images in the wildfire detection model. The model generated 10 false negatives (missed wildfires) and 10 false positives (incorrect wildfire classification) according to the confusion matrix. These mistakes might be ascribed to elements including Cloud Cover and Smoke Obstruction, Nightfall and Low-Light Conditions.

The training vs. validation loss graph is shown in "Fig. 8". The training accuracy steadily increases, exceeding 95%, while validation loss fluctuates. With training accuracy above validation accuracy, the graph showing training against validation accuracy shows growing accuracy in "Fig. 7".

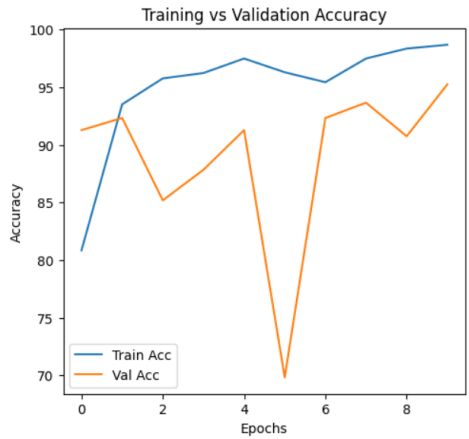

Fig. 7 Training vs Validation Accuracy Graph

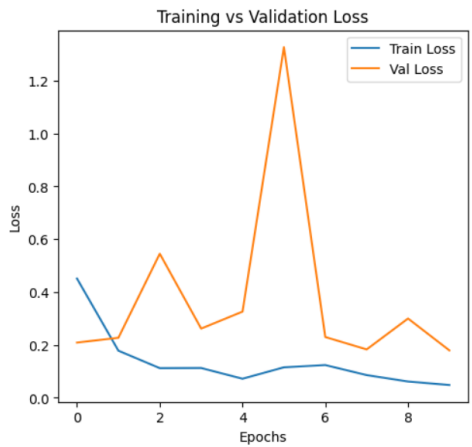

Fig. 8 Training vs Validation Loss Graph

The model used 1089.92 MB of memory shown in "Fig. 9", demonstrating efficient GPU usage. Reflecting computational challenge, the backward pass required 0.0243 seconds; the forward pass took 0.0072 seconds for each batch. Each epoch took 406.46 seconds, and inference per batch was 0.0068 seconds, as shown in "Fig. 10" indicating optimized performance for real-time wildfire detection applications.

Memory Used: 1089.92 MB

Fig. 9 Memory Used

Forward Pass Time per Batch: 0.007202 seconds
Backward Pass Time per Batch: 0.024324 seconds
Training Time per Epoch: 406.46 seconds
Inference Time per Batch: 0.006795 seconds

Fig. 10 Time Complexity

Using deep learning approaches, the table shows several wildfire detection methods especially Vision Transformers (ViTs) and hybrid models. It contrasts research employing sequential fire images, UAV feeds, and satellite imaging among other datasets. Using ViT-based classification on high-resolution wildfire data, our approach obtained an accuracy of 96.10%.

TABLE 1.  Comparison of Proposed and Existing Algorithms Percentage of Accuracy

| Authors | Methods | Datasets | Result |
| --- | --- | --- | --- |
| Wang (2023) | Vision Transformer (ViT) for wildfire detection | Satellite imagery (infrared & multi-spectral) | 92.5% |
| Lopez (2023) | Hybrid CNN-Transformer for wildfire prediction | Time-series satellite data & meteorological parameters | 91.8% |
| Zhang and Chen (2022) | ViT-based wildfire risk assessment | Satellite imagery & topographical data | 93.2% |
| Ghosh (2021) | ViT + RNN for wildfire spread prediction | Wildfire Dataset (sequential fire progression images) | 92.7% |
| Roy (2020) | Transformer-based real-time wildfire detection | UAV & satellite feeds | 92.9% |
| Proposed Method (2025) | ViT-based classification for wildfire detection | The Wildfire Dataset | 96.10% |

TABLE 2.  Comparison of Proposed and Other Algorithms Percentage of Accuracy

| Model | Accuracy |
| --- | --- |
| ResNet-50 | 88.5% |
| VGG-16 | 85.3% |
| YOLOv5 | 87.2% |
| AlexNet | 83.9% |
| Vision Transformer | 96.10% |

V. CONCLUSIONS AND FUTURE WORK

This study proposes the use of Vision Transformers (ViTs) to enhance wildfire detection accuracy. By using self-attention techniques to capture long-range dependencies, ViTs efficiently separate between wildfire and non-wildfire images. Employing a well-structured data preparation process, the model was trained using 2,700 high-resolution aerial and ground images from the Wildfire Dataset, therefore guaranteeing consistency in image dimensions and enhanced convergence. Demonstrating better classification ability, the ViT-Base-Patch16-224 model exceeded conventional CNN-based designs including ResNet, VGG-16, YOLO, and AlexNet as shown in "Table. 2". The model showed a constant drop in loss during training, from 0.4510 in the first epoch to 0.0481 in the last epoch, therefore verifying the effectiveness of the Adam optimizer. With few false positives (10) and false negatives (10), the confusion matrix emphasizes the model's robustness and guarantees great detection dependability. The model also obtained an F1-score of 0.97, therefore confirming its efficiency in precisely spotting wildfires and lowering false alarms. Our ViT-based wildfire detection system exceeds current techniques with an enhanced ultimate accuracy of 96.10%, therefore highlighting the possibilities of transformer-based architectures for sophisticated environmental image categorization and real-time wildfire monitoring.

Minimizing false negatives in wildfire detection is critical, as undetected fires can escalate rapidly, leading to catastrophic consequences. A missed wildfire can inflict major environmental harm, loss of human and animal life, and general ruin. Delays in fire containment also endanger firefighters and raise suppressing expenses. Early, correct identification guarantees a quick response, so stopping the uncontrollable spreading of flames. Our model's recall of 0.98 minimizes false negatives, thereby enhancing the response efficacy. Using Vision Transformers (ViTs) and real-time data helps us greatly lower the chance of undetectable wildfires, hence improving catastrophe readiness and mitigating techniques.

Future developments in wildfire detection should center on improving model performance, real-time deployment, and dataset extension. Using a mix of hybrid ViT-CNN designs could improve the accuracy of feature extraction and classification, making detection more reliable in a wider range of settings. Using UAV and satellite-based surveillance can also help to follow wildfires in real-time, therefore enhancing the effectiveness of emergency response.

Including multi-spectral, infrared, and night-time images in the Wildfire Dataset will improve the model's capacity to identify fires under diverse circumstances. Synthetic data augmentation methods are also capable of resolving class imbalance problems. Moreover, optimizing ViTs for real-time edge deployment and reduced computing cost helps to enable quicker decisions.

Lastly, adding techniques for explainability like attention heatmaps would make things easier to understand, which would boost trust in AI-based systems that find wildfires. Future research could investigate self-supervised learning techniques to improve wildfire prediction without depending on large labeled datasets, hence optimizing AI-driven disaster management utilizing scalability and efficiency.

ACKNOWLEDGMENT

This research is funded by NASA Connecticut Space Grant Consortium under Faculty Research Grant, PTE Federal Award No.: 80NSSC20M0129, grant period: 7/1/2024-05/31/2025. The authors are grateful for the support from NASA CT Space Grant Consortium.